# Propagation of 2-Monotone Lower Probabilities on an Undirected Graph

**Lonnie Chrisman**
School of Computer Science
Carnegie Mellon University
Pittsburgh, PA 15217
chrisman@cs.cmu.edu

## Abstract

Lower and upper probabilities, also known as *Choquet capacities*, are widely used as a convenient representation for sets of probability distributions. This paper presents a graphical decomposition and exact propagation algorithm for computing marginal posteriors of 2-monotone lower probabilities (equivalently, 2-alternating upper probabilities).

## 1 Introduction

Let $(\Omega, \mathcal{F})$ be a probability space, and $\mathcal{P}$ a non-empty set of probability distributions on that space. The functions

$$\underline{P}(A) = \inf_{P \in \mathcal{P}} P(A) \qquad \overline{P}(A) = \sup_{P \in \mathcal{P}} P(A) \qquad (1)$$

defined for any $A \in \mathcal{F}$, are lower and upper probability envelopes.

A number of uses have been suggested for lower probabilities, and their use is rapidly increasing. Some feel that the use of a single exact distribution in Bayesian-style inference fails to satisfactorily distinguish between uncertainty and ignorance or between certainty and confidence, and therefore a more general representation such as lower probability functions may be a superior representation of belief [32, 44]. Lower probabilities may also arise from incomplete or partial elicitation, such as when insufficient knowledge is available, or when it is too time consuming to obtain the necessary knowledge to warrant the precision inherent in exact probabilities [16, 20]. Lower probabilities are also useful for studying sensitivity and robustness in probabilistic inference [3, 46, 49], and they can be used to weigh computation effort against modeling precision [11]. They arise in group decision problems [39] and in axiomatic approaches to uncertainty when the axioms of probability are weakened [18, 46]. They arise when determining constraints on probabilities given only the probabilities on a finite set of other events [35]. Finally, they may result from the abstraction of more detailed probabilistic models [8, 10, 21].

In recent years, graphical decompositions of probability distributions have found widespread interest and application [23, 36]. These representations not only admit a concise and structured specification of a joint probability distribution, but also allow marginal posterior probabilities to be efficiently computed by taking advantage of the graphical structure [12, 25, 31, 36]. Analogous decompositions for lower probabilities may present similar opportunities for the many previously cited applications areas. Unfortunately, lower probabilities seem to be rather resistant to propagation. For example, if they are updated incrementally as evidence arrives, as is often done in Bayesian applications of pure probability, the resulting bounds depend on the order that evidence arrives, and are not equivalent to the bounds obtained by updating the original prior with all the evidence in a single step [9, 22, 37]. Fundamentally, the representation looses information during the updating process. Any direct attempt at propagation would almost certainly experience the same losses of information at each propagation step, and could invariably push the representation to either vacuous or inconsistent probability bounds.

A number of previous works have developed propagation algorithms for various representations of convex sets of probabilities. Notably, [4, 5] developed an algorithm, based on the axioms of [43], to propagate convex sets of probabilities represented by convex finite-sided polytopes in the probability simplex. Their polytopes are a generalization of the lower probabilities considered here; however, there are good reasons why one might prefer to deal with the lower probability representation directly rather than the more general convex probability formalism. For example, lower probabilities are more natural, refer only to the basic events in the probability space (rather than surfaces in the probability simplex), and have potential to be much more computationally tractable. Importantly, the propagation of polytopes can cause the number of sides to increase multiplicatively with each propagation step ([44]). [16] and [44] developed algorithms to propagate probability bounds in graphical structures. Their bounds, being only over individual elements of the probability space, are special cases of lower prob-ability and can be substantially less informative, so the interest in the more expressive lower probability remains. Logics for reasoning with constraints on the probability of an incomplete set of events have been used by [1, 35, 38, 45], and indeed some even call the inference process "propagation". The problem addressed by these and the use of the term propagation are considerably different than those in this paper since our initial bounds are prior lower probabilities as opposed to absolute constraints. When combined with knowledge of other propositions, our local prior bounds are subject to revision, so they are not constraints in that sense. A Dempster-Shafer belief function is syntactically a special case of a 2-monotone lower probability, and a number of papers have developed decompositions and propagation algorithms for belief functions, e.g., [29, 42, 50]. In these examples, however, the belief function is given an evidential interpretation [22, 41], rather than a lower probabilistic one. Propagation of second-order distributions has been considered by [33] and [34]. To date, the author is not aware of any previously published algorithm to propagate lower probability bounds directly.

This paper introduces a decomposition and propagation algorithm for the lower probability representation. It overcomes the apparent resistance of lower probability to propagation by utilizing the observation of [9] that the problems with lower probability updating can be alleviated by using a different and more informative internal representation for the bounds. [9] shows that the more informative internal representation can actually be substantially more efficient. The new propagation algorithm allows evidence to be locally and incrementally incorporated, and marginal posterior lower probabilities to be computed via propagation.

## 2 Lower Probability

Let $(\Omega, \mathcal{F} = 2^\Omega)$ be a finite probability space, and $\underline{P} : \mathcal{F} \longrightarrow [0,1]$ be a set function on this space. $\underline{P}$ is called a *2-monotone lower probability* (or *2-monotone Choquet capacity* [7]) when for any $A, B \in \mathcal{F}$ the following hold:

1. $\underline{P}(\emptyset) = 0$, $\underline{P}(\Omega) = 1$
2. $\underline{P}(A) + \underline{P}(B) \leq \underline{P}(A \cup B) + \underline{P}(A \cap B)$

Not every probability envelope (as defined by (1)) is 2-monotone, but most applications restrict attention to 2-monotone representations since it is the weakest property that readily admits simple closed-form manipulations.

A dual set function, called a *2-alternating upper probability* (or 2-alternating Choquet capacity), is given by $\overline{P}(A) = 1 - \underline{P}(\bar{A})$, where $\bar{A} = \Omega - A$ denotes the complement of $A$. It follows that for any $A, B \in \mathcal{F}$,

1. $\overline{P}(\emptyset) = 0$, $\overline{P}(\Omega) = 1$
2. $\overline{P}(A) + \overline{P}(B) \geq \overline{P}(A \cup B) + \overline{P}(A \cap B)$
3. $\underline{P}(A) \leq \overline{P}(A)$

A probability distribution $P$ on $(\Omega, \mathcal{F})$ is said to be *consistent* with $\underline{P}$ if for all $A \in \mathcal{F}$, $\underline{P}(A) \leq P(A)$, or equivalently, $P(A) \leq \overline{P}(A)$. Denote the set of all distributions consistent with $\underline{P}$ by $\mathcal{P}(\underline{P})$. Every two-monotone lower probability has at least one consistent probability distribution. Lower probabilities are natural representations for convex sets of distributions — namely, $\underline{P}$ represents the set of all distributions on $(\Omega, \mathcal{F})$ consistent with $\underline{P}$. Note that there are many different convex sets of distributions with bounds given by $\underline{P}$, but when $\underline{P}$ is specified, it normally is assumed to represent only the maximal such set.

Let $\underline{P}(A|E) = \inf\{P(A|E) : P \in \mathcal{P}(\underline{P}), P(E) > 0\}$. When $\underline{P}$ is 2-monotone and $\underline{P}(E) > 0$, it is well-known [9, 14, 15, 48] that

$$\underline{P}(A|E) = \frac{\underline{P}(A \cap E)}{\underline{P}(A \cap E) + \overline{P}(\bar{A} \cap E)}$$

$$\overline{P}(A|E) = \frac{\overline{P}(A \cap E)}{\overline{P}(A \cap E) + \underline{P}(\bar{A} \cap E)} \qquad (2)$$

When $0 = \underline{P}(E) < \overline{P}(E)$, then $\underline{P}(A|E) = 1$ whenever $A \subset E$ and is zero otherwise ([9]). Equation (2) also provides valid but non-exact bounds when $\underline{P}$ is not 2-monotone, and is the only instance in this paper where 2-monotonicity is used.

The *Möbius transform* of a lower probability function is defined by ([40, pg. 39])

$$m(A) = \mathsf{M}[\underline{P}](A) = \sum_{B \subset A} (-1)^{|A-B|} \underline{P}(B) \qquad (3)$$

The summation in (3) is taken over all sets $B \in \mathcal{F}$ such that $B \subset A$, but in all the summations that follow, we suppress $B \in \mathcal{F}$ from the notation here for convenience. If $m(A)$ is non-negative on all sets $A \in \mathcal{F}$, $\underline{P}$ is said to be *infinitely-monotone*, and is also often referred to as a *belief function*. The Möbius transform is information preserving, such that the original function $\underline{P}$ can be recovered from $m$ using the *inverse Möbius transform* given by ([40, Lemma 2.3], [6, Appendix]):

$$\underline{P}(A) = \mathsf{M}^{-1}[m](A) = \sum_{B \subset A} m(B)$$

The *commonality transform* of an upper probability function is defined by ([40, pg. 44]):

$$Q(A) = \mathsf{Q}[\overline{P}](A) = -\sum_{B \subset A} (-1)^{|B|} \overline{P}(B) \qquad (4)$$

when $A \neq \emptyset$ and $Q(\emptyset) = 1$. This transform is also information preserving, so that the original $\overline{P}$ can be recovered from $Q$ using the *inverse commonality transform* given by ([40, Theorem 2.6])

$$\overline{P}(A) = \mathsf{Q}^{-1}[Q](A) = 1 - \sum_{B \subset A} (-1)^{|B|} Q(B)$$

**Proposition 1** *Let $A, E \in \mathcal{F}$. The following hold.*

$$\underline{P}(A \cap E) = \mathsf{M}^{-1}[m'](A)$$

$$m'(A) = \begin{cases} m(A) & \text{if } A \subset E, \\ 0 & \text{otherwise} \end{cases} \quad (5a)$$

$$\overline{P}(A \cap E) = \mathsf{Q}^{-1}[Q'](A)$$

$$Q'(A) = \begin{cases} Q(A) & \text{if } A \subset E, \\ 0 & \text{otherwise} \end{cases} \quad (5b)$$

*where $m = \mathsf{M}[\underline{P}]$ and $Q = \mathsf{Q}[\overline{P}]$.*

Proposition 1 allows evidence to be incorporated into the representation. Both $m'$ and $Q'$ in (5) can be incrementally updated without a loss of information ([9]). In other words, if $E_1$ is learned, and then $E_2$ is later learned, $E_1$ can be incorporated first, and $m$ replaced by $m'$. Then when $E_2$ is learned, $E_2$ can be incorporated into the new representation. The final representation is identical to that obtained by incorporating $E_1 \cap E_2$ into the original belief in one step. Thus, by maintaining $m'$ and $Q'$ internally, $\underline{P}(A|E)$ and $\overline{P}(A|E)$ can be incrementally updated and at any time obtained from (2).

## 3 Joint Lower Probability

Let $\mathbf{V} = \{\mathbf{x}_1, ..., \mathbf{x}_k\}$ be a set of variables taking on possible values from $\Omega_1, ..., \Omega_k$. A joint assignment, $x = \times_{i=i..k} x_i$, takes on a value from $\Omega = \Omega_1 \times ... \times \Omega_k$. Again, we assume $\Omega_i$ is finite, $\mathcal{F} = 2^\Omega$. If $\mathbf{A} \subset \mathbf{V}$ is a subset of variables ($\mathbf{A} = \{\mathbf{x}_{i_1}, ..., \mathbf{x}_{i_\ell}\}$), then $x_\mathbf{A}$ denotes $x_{i_1} x_{i_2} ... x_{i_\ell}$, and $\Omega_\mathbf{A} = \Omega_{i_1} \times ... \times \Omega_{i_\ell}$. Note that boldface capital letters denote subsets of variables, while non-bold capital letters denote subsets of $\Omega$, and non-bold small letter denote elements of $\Omega$. When $\mathbf{A} \subset \mathbf{V}$ and $B \in \mathcal{F}$, $B_\mathbf{A}$ denotes $\{x_\mathbf{A} : x \in B\}$; therefore, $B_\mathbf{A} \in \mathcal{F}_\mathbf{A} = 2^{\Omega_\mathbf{A}}$.

Since $(\Omega_\mathbf{A}, \mathcal{F}_\mathbf{A})$ is a probability space, a lower probability, $\underline{P}_\mathbf{A}$, can be specified over this space as well. The subscript on $\underline{P}_\mathbf{A}$ is used for notational clarity to indicate the underlying space if it is not $(\Omega, \mathcal{F})$ (i.e., no subscript is equivalent to a subscript $\mathbf{V}$). Subscripts are similarly used on $m_\mathbf{A}$ and $Q_\mathbf{A}$. These subscripts are not operators, they simply distinguish different functions in a way that always makes it clear what the underlying space is. When $A \in \mathcal{F}_\mathbf{A}$ is a set and $\mathbf{B}$ a subset of variables, $\mathbf{B} \supset \mathbf{A}$, then $A^{\uparrow \mathbf{B}} = A \times \{\Omega_{\mathbf{B} \setminus \mathbf{A}}\}$, and $A^\uparrow = A^{\uparrow \mathbf{V}}$

Given a joint lower probability in a propagation framework, one is normally interested in the marginal lower probability over a small subset of variables. The propagation framework allows these marginals to be computed efficiently. The functions $Loc^m(m, \mathbf{A})$ and $Loc^Q(Q, \mathbf{A})$, defined below, return the marginal of $m$ or $Q$ *loc*alized to the given variables in $\mathbf{A}$. These are defined by

$$Loc^m(m_\mathbf{B}, \mathbf{A})(A) = \sum_{\substack{B \in \mathcal{F}_\mathbf{B} \\ B_\mathbf{A} = A}} m_\mathbf{B}(B) \quad (6a)$$

$$Loc^Q(Q_\mathbf{B}, \mathbf{A})(A) = (-1)^{|A|} \sum_{\substack{B \in \mathcal{F}_\mathbf{B} \\ B_\mathbf{A} = A}} (-1)^{|B|} Q_\mathbf{B}(B) \quad (6b)$$

where $m_\mathbf{B}$ and $Q_\mathbf{B}$ are Möbius or commonality assignments over $(\Omega_\mathbf{B}, \mathcal{F}_\mathbf{B})$. Both of (6a) and (6b) are generalizations of the standard notion of marginalization of a point probability distribution. It is important that when $\mathbf{A} \subset \mathbf{B}$, then $Loc^m(Loc^m(m, \mathbf{B}), \mathbf{A}) = Loc^m(m, \mathbf{A})$ and $Loc^Q(Loc^Q(Q, \mathbf{B}), \mathbf{A}) = Loc^Q(Q, \mathbf{A})$.

The following fundamental theorem of marginal lower probabilities states that marginal lower probabilities are related to joint lower probabilities in the manner that one would intuitively expect.

**Theorem 1** *Denote $m_\mathbf{B} = \mathsf{M}[\underline{P}_\mathbf{B}]$ and $Q_\mathbf{B} = \mathsf{Q}[\overline{P}_\mathbf{B}]$. Let $\mathbf{A} \subset \mathbf{B}$ and*

$$m_\mathbf{A}(A) = Loc^m(m_\mathbf{B}, \mathbf{A})(A)$$

$$Q_\mathbf{A}(A) = Loc^Q(Q_\mathbf{B}, \mathbf{A})(A)$$

*Then $\mathsf{M}^{-1}[m_\mathbf{A}](A) = \underline{P}_\mathbf{B}(A^{\uparrow \mathbf{B}})$ and $\mathsf{Q}^{-1}[Q_\mathbf{A}](A) = \overline{P}_\mathbf{B}(A^{\uparrow \mathbf{B}})$ for all $A \in \mathcal{F}_\mathbf{A}$.*

For example, the marginal bounds as defined by (6) for an event $A \in \mathcal{F}_\mathbf{A}$ are just $[\underline{P}(A^\uparrow), \overline{P}(A^\uparrow)]$. The combination of (2) with Theorem 1 and Proposition 1 provides the basis for computing marginal posterior lower probabilities. Suppose $\mathbf{A}$ is a subset of variables, and one wishes to compute the marginal posterior $\underline{P}(A^\uparrow|E)$ where $A \in \mathcal{F}_\mathbf{A}$ and $E \in \mathcal{F}$. This is obtained from

$$\underline{P}(A^\uparrow|E) = \frac{\mathsf{M}^{-1}[Loc^m(m', \mathbf{A})](A)}{\mathsf{M}^{-1}[Loc^m(m', \mathbf{A})](A) + \mathsf{Q}^{-1}[Loc^Q(Q', \mathbf{A})](\bar{A})} \quad (7)$$

where $m'$ and $Q'$ are given by (5a) and (5b). The same bound can be obtained (when $\underline{P}$ is 2-monotone) by applying Bayes's rule to all distributions consistent with $\underline{P}$, marginalizing all of them to $\mathbf{A}$, and taking the lower bound.

## 4 Graphical Decomposition

Let $\mathcal{G} = (\mathbf{V}, E)$ be an undirected graph with vertices $\mathbf{V}$ and edges $E \subset \{\{\alpha, \beta\} : \alpha, \beta \in \mathbf{V}, \alpha \neq \beta\}$. The vertices of our graphs correspond to the random variables $\mathbf{V}$ above, hence the dual use of $\mathbf{V}$. A *path* of length $L$ from $\alpha_0$ to $\alpha_L$ is a sequence of at least two vertices, $\alpha_0, \alpha_1, ..., \alpha_L$, such that $\{\alpha_i, \alpha_{i+1}\} \in E$. A *cycle* is a path with $\alpha_0 = \alpha_L$. A subset of vertices, $\mathbf{S}$, is said to *separate* $\mathbf{A}$ from $\mathbf{B}$ when all paths from any node of **A** to any node of **B** contain a node in **S**. If **A** is a subset of vertices, the graph $\mathcal{G}_\mathbf{A}$ induced by **A** is the subgraph $\mathcal{G}_\mathbf{A} = (\mathbf{A}, E_\mathbf{A})$ where $E_\mathbf{A}$ is the set of edges in $E$ with both endpoints in **A** (i.e., $E_\mathbf{A} = \{\{\alpha, \beta\} : \alpha, \beta \in \mathbf{A}\} \cap E$). A subset of vertices, **A**, is called *complete* when all pairs of vertices in **A** are connected. If **A** complete and is not a subset of a larger complete set of vertices, then **A** is called a *clique*. The set of all cliques in $\mathcal{G}$ is denoted by $\mathcal{C}$.

A pair of vertex subsets, $(\mathbf{A}, \mathbf{B})$, *decomposes* $\mathcal{G}$ when $\mathbf{V} = \mathbf{A} \cup \mathbf{B}$, $\mathbf{A} \cap \mathbf{B}$ is complete, and $\mathbf{A} \cap \mathbf{B}$ separates **A** from **B**. The decomposition is called *proper* if $\mathbf{A}, \mathbf{B} \neq \mathbf{V}$. A graph $\mathcal{G}$ is *decomposable* when it is complete, or if there exists a proper decomposition $(\mathbf{A}, \mathbf{B})$ into decomposable subgraphs $\mathcal{G}_\mathbf{A}$ and $\mathcal{G}_\mathbf{B}$. A well known graph theoretic result (e.g., [19]) is that a graph is decomposable if and only if it is *triangulated* (also called *chordal*), that is, if all cycles of length $L \geq 4$ contain a short-circuiting edge (a *chord*) between two non-consecutive vertices in the cycle. Any graph can be converted to a *triangulated* graph by adding edges, but finding the optimal triangulation is in most cases $NP$-hard ([2]). Heuristics for triangulation are often effective ([27]). The above is standard graph-theoretic terminology. For further reference see [19].

It is also convenient to introduce some additional terminology. If $\mathcal{G}$ is a graph with cliques $\mathcal{C}$, and $A \in \mathcal{F}$ is a set, the *rectangularization* of $A$ with respect to $\mathcal{G}$ is

$$^\square A = \bigcap_{\mathbf{C} \in \mathcal{C}} A_\mathbf{C}^\dagger$$

$A$ is a *rectangular set* (or just *rectangle*) on $\mathcal{G}$ when $^\square A = A$. It is always the case that $A \subset {^\square A}$, and furthermore, $^\square A$ is the smallest rectangle containing $A$. Denote the set of all rectangular sets on $\mathcal{G}$ by $\mathcal{R}$, and the set of all rectangular sets on subgraph $\mathcal{G}_\mathbf{A}$ by $\mathcal{R}_\mathbf{A}$. We say a lower probability, $\underline{P}$, has a *rectangular core* on $\mathcal{G}$ if $\mathsf{M}[\underline{P}](A) = 0$ whenever $A \notin \mathcal{R}$, or equivalently, when $\mathbb{Q}[\overline{P}](A) = \mathbb{Q}[\overline{P}]({^\square A})$ for all $A \in \mathcal{F}$.

**Definition 1** *Let $(\mathcal{G}^m, \mathcal{G}^Q)$ be a pair of decomposable graphs, and let $\mathcal{R}^m$ and $\mathcal{R}^Q$ be the rectangular subsets on $\mathcal{G}^m$ and $\mathcal{G}^Q$ respectively. We say that a lower probability $\underline{P}$ is Markov with respect to $(\mathcal{G}^m, \mathcal{G}^Q)$ when for any decomposition $(\mathbf{A}, \mathbf{B})$ of $\mathcal{G}^m$ and any decomposition $(\mathbf{C}, \mathbf{D})$ of $\mathcal{G}^Q$,*

$$\mathsf{M}[\underline{P}](A) = \frac{Loc^m(\mathsf{M}[\underline{P}], \mathbf{A})(A_\mathbf{A}) \cdot Loc^m(\mathsf{M}[\underline{P}], \mathbf{B})(A_\mathbf{B})}{Loc^m(\mathsf{M}[\underline{P}], \mathbf{A} \cap \mathbf{B})(A_{\mathbf{A} \cap \mathbf{B}})}$$

*when $A \in \mathcal{R}^m$, $\mathsf{M}[\underline{P}](A) = 0$ when $A \notin \mathcal{R}$, and*

$$\mathbb{Q}[\overline{P}](A) = \frac{Loc^Q(\mathbb{Q}[\overline{P}], \mathbf{C})({^\square A}_\mathbf{C}) \cdot Loc^Q(\mathbb{Q}[\overline{P}], \mathbf{D})({^\square A}_\mathbf{D})}{Loc^Q(\mathbb{Q}[\overline{P}], \mathbf{C} \cap \mathbf{D})({^\square A}_{\mathbf{C} \cap \mathbf{D}})}$$

In other words, $m$ and $Q$ must both be individually Markov and $\underline{P}$ must have a rectangular core on both $\mathcal{G}^m$ and $\mathcal{G}^Q$. The extra requirement that $\underline{P}$ have a rectangular core is a technical detail that appears when generalizing decomposability to non-additive set functions. Without this restriction, one could place mass-assignments on sets that are not properly discerned by existing cliques in the graph. Not only would it be unnatural to consider a lower probability with such assignments to be Markov, such assignments create technical inconsistencies.

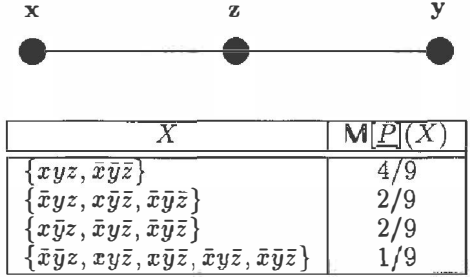

| $X$ | $\mathsf{M}[\underline{P}](X)$ |
|---|---|
| $\{xyz, \bar{x}\bar{y}\bar{z}\}$ | 4/9 |
| $\{\bar{x}yz, x\bar{y}\bar{z}, \bar{x}\bar{y}\bar{z}\}$ | 2/9 |
| $\{x\bar{y}z, \bar{x}y\bar{z}, \bar{x}\bar{y}\bar{z}\}$ | 2/9 |
| $\{\bar{x}\bar{y}z, xy\bar{z}, x\bar{y}\bar{z}, \bar{x}y\bar{z}, \bar{x}\bar{y}\bar{z}\}$ | 1/9 |

Figure 1: Two (identical) sensors, $x$ and $y$, are set up to detect an earthquake, $z$. Each sensor functions correctly in 2/3 of all cases. In the remaining cases, it does not sound when there is an earthquake, but may or may not function correctly when there is no earthquake. So that no further distributional assumptions are made, the joint reliability of the sensors is modeled by a lower probability, factorized according to the graph shown above. For conciseness, only non-zero Möbius assignments are shown. The joint lower probability is Markov — both $m$ and $Q$ factor on the graph.

It may be convenient or appropriate to enforce $\mathcal{G}^m = \mathcal{G}^Q$, so that there is only one graph being considered, but this is not required and not doing so may allow additional flexibility. For example, each of the two Markov conditions might be achieved by triangulating $\mathcal{G}^m$ and $\mathcal{G}^Q$ differently. However, because the case of $\mathcal{G}^m = \mathcal{G}^Q$ is of significant interest, it is informative to consider the conditions in which such a Markov decomposition is possible. Figure 1 shows an example of a Markov lower probability, establishing that interesting Markov lower probabilities do exist (of course, a pure probability decomposition satisfies the conditions as well). The following establishes conditions in which it is possible for $\mathcal{G}^m$ and $\mathcal{G}^Q$ to be the same graph.

**Theorem 2** *Let $m = \mathsf{M}[\underline{P}]$. If for any decomposition $(\mathbf{A}, \mathbf{B})$ of $\mathcal{G}$ and any set $X \in \mathcal{R}$ with $m(X) \neq 0$, there is a unique pair $A \in \mathcal{F}_\mathbf{A}$, $B \in \mathcal{F}_\mathbf{B}$, $X = A^\dagger \cap B^\dagger$, such that $Loc^m(m, \mathbf{A})(A) \neq 0$ and $Loc^m(m, \mathbf{B})(B) \neq 0$, then $\mathsf{M}[\underline{P}]$ is Markov on $\mathcal{G}$ if any only if $\mathbb{Q}[\overline{P}]$ is Markov on the same graph $\mathcal{G}$. When $\underline{P}$ is infinitely-monotone, then this is also a necessary condition for $\mathsf{M}[\underline{P}]$ and $\mathbb{Q}[\overline{P}]$ to be Markov on the same graph.*

**Corollary 1** *If for every decomposition $(\mathbf{A}, \mathbf{B})$ of $\mathcal{G}$, $\{X_{\mathbf{A} \cap \mathbf{B}} : \mathsf{M}[\underline{P}](X) \neq 0\}$ partitions $\Omega_{\mathbf{A} \cap \mathbf{B}}$, then $\mathsf{M}[\underline{P}]$ is Markov on $\mathcal{G}$ if and only if $\mathbb{Q}[\overline{P}]$ is Markov on the same graph $\mathcal{G}$.*

For any decomposable graph $\mathcal{G}$, one can efficiently identify a tree, $\mathcal{J} = (\mathcal{C}, \mathcal{S})$, called a *junction tree* ([24]), with the following properties

1. Each node of $\mathcal{J}$ contains a subset of nodes of $\mathcal{G}$, corresponding to a clique of $\mathcal{G}$.

2. The intersection of all node subsets on a path of $\mathcal{J}$ is equal to the intersect of the node subsets of the path's endpoints.

Let $\mathcal{J}^m = (\mathcal{C}^m, \mathcal{S}^m)$ and $\mathcal{J}^Q = (\mathcal{C}^Q, \mathcal{S}^Q)$ be junction trees for $\mathcal{G}^m$ and $\mathcal{G}^Q$ respectively. A *clique potential*, $\phi_\mathbf{C} : \mathcal{F}_\mathbf{C} \longrightarrow \Re$, is attached to each node of $\mathcal{J}^m$ and $\mathcal{J}^Q$, and a *separator potential*, $\phi_\mathbf{S} : \mathcal{F}_\mathbf{S} : \mathcal{F}_\mathbf{S} \longrightarrow \Re$, is attached to each edge of $\mathcal{J}^m$ and $\mathcal{J}^Q$, where the edge is between nodes $\mathbf{A}$ and $\mathbf{B}$ and $\mathbf{S} = \mathbf{A} \cap \mathbf{B}$. Denote these $\phi^m$ and $\phi^Q$. These potentials are initialized so that for any $A \in \mathcal{F}$

$$\mathsf{M}[\underline{P}](A) = \begin{cases} \dfrac{\prod_{\mathbf{C} \in \mathcal{C}^m} \phi_\mathbf{C}^m(A_\mathbf{C})}{\prod_{\mathbf{S} \in \mathcal{S}^m} \phi_\mathbf{S}^m(A_\mathbf{S})} & A \in \mathcal{R} \\ 0 & \text{otherwise} \end{cases}$$

$$\mathsf{Q}[\overline{P}](A) = \dfrac{\prod_{\mathbf{C} \in \mathcal{C}^Q} \phi_\mathbf{C}^Q(A_\mathbf{C})}{\prod_{\mathbf{S} \in \mathcal{S}^Q} \phi_\mathbf{S}^Q(A_\mathbf{S})} \qquad (8)$$

It can be said that the potentials encode the joint prior $m$ and $Q$ functions. The basis for this initialization depends on the application and is not considered here. Nevertheless, a few comments about initialization are in order.

It is important to identify frameworks in which joint lower probabilities can be constructed out of convenient bits and pieces. For example, Bayesian networks provide a means for constructing joint probability distributions out of local conditional probabilities. As is the case with probabilities, we envision there being many possible frameworks that might provide convenient ways of constructing joint lower probabilities from components. The only requirement is that the joint probability be expressed in the product form of (8). Probabilistic Markov field theory provides the foundation for computation in many frameworks, including Bayesian networks ([31]), Markov networks ([36]), chain graphs ([17]), influence diagrams ([26]), Markov processes and temporal probabilistic networks ([28]), etc. In the same way, the product representation here may serve as the underlying computational foundation for a variety of application frameworks. The particular way in which components are specified may depend on the particular goals of the application, interpretation of the lower probabilities, desired properties of the representation, and other considerations. The bare framework of this paper can be used directly if the components in (8) can be assessed directly. However, it is clear that the development of more natural frameworks is an area of research in critical need of further attention.

### 4.1 Propagation

We say potentials $\phi_\mathbf{A}^m$ and $\phi_\mathbf{B}^m$, $\mathbf{A}, \mathbf{B} \in \mathcal{C}^m$, are *consistent* when $Loc^m(\phi_\mathbf{A}^m, \mathbf{A} \cap \mathbf{B}) = Loc^m(\phi_\mathbf{B}^m, \mathbf{A} \cap \mathbf{B})$. Similarly, $\phi_\mathbf{A}^Q$ and $\phi_\mathbf{B}^Q$, $\mathbf{A}, \mathbf{B} \in \mathcal{C}^Q$, are consistent when $Loc^Q(\phi_\mathbf{A}^Q, \mathbf{A} \cap \mathbf{B}) = Loc^Q(\phi_\mathbf{B}^Q, \mathbf{A} \cap \mathbf{B})$.

**Theorem 3 (Uniqueness)** *(I) Suppose for each node $\mathbf{C} \in \mathcal{C}^m$ of $\mathcal{G}^m$, a potential $\phi_\mathbf{C}^m$ is specified, and that these potentials are pairwise consistent. Then there is a unique Markov Möbius assignment, $m$, having $Loc^m(m, \mathbf{C}) = \phi_\mathbf{C}^m$.*

*(II) Suppose for each node $\mathbf{C} \in \mathcal{C}^Q$ of $\mathcal{G}^Q$, a potential $\phi_\mathbf{C}^Q$ is specified, and that these potentials are pairwise consistent. Then there is a unique Markov commonality assignment, $Q$, having $Loc^Q(Q, \mathbf{C}) = \phi_\mathbf{C}^Q$.*

The initial potentials are not, in general, pairwise consistent. The propagation algorithm leaves the joint potential unaltered, but changes the local potentials so that they are pairwise consistent, and therefore by Theorem 3, the marginals can be directly read off from the local potentials.

The propagation of $\phi^m$ and $\phi^Q$ can each be done separately — there is no interaction between these during the propagation. The propagation of each occurs in the same fashion. Here $\phi$ denotes either $\phi^m$ or $\phi^Q$, $Loc$ either $Loc^m$ or $Loc^Q$, and so on.

A full propagation proceeds as follows (done for both junction trees $\mathcal{J}^m$ and $\mathcal{J}^Q$). Any node of $\mathcal{J}$ is chosen as the root. Let $d$ be the maximum distance in $\mathcal{J}$ between the root and any other node. First, during the *collect evidence* stage, each node at depth $d$ propagates potential information to its neighbor at depth $d-1$. Then each node at depth $d-1$ propagates to its neighbor at depth $d-2$, and so on until the root's neighbors have propagated to the root. Second, during the *distribute evidence* stage, the root propagates information to each of its neighbors, then they propagate to each of their neighbors at depth 2, and so on down to depth $d$.

A propagation step from node $\mathbf{A} \in \mathcal{C}$ to node $\mathbf{B} \in \mathcal{C}$ occurs as follows (let $\mathbf{S} = \mathbf{A} \cap \mathbf{B}$):

$$\phi_\mathbf{S}' = Loc(\phi_\mathbf{A}, \mathbf{S})$$
$$\phi_\mathbf{B}'(B) = \phi_\mathbf{B}(B) \phi_\mathbf{S}'(B_\mathbf{S}) / \phi_\mathbf{S}(B_\mathbf{S}), \qquad B \in \mathcal{F}_\mathbf{B}$$

Where we take $0/0 = 0$. Then $\phi_\mathbf{S}$ and $\phi_\mathbf{B}$ are replaced by $\phi_\mathbf{S}'$ and $\phi_\mathbf{B}'$. See e.g., [12, 25, 31].

If either of the initial graphs, $(\mathcal{G}^m, \mathcal{G}^Q)$, has disconnected components, it is essential that a single connected junction tree be used for each $\mathcal{J}^m$ and $\mathcal{J}^Q$ that includes all the disconnected components. This can be accomplished by including an artificial node in the junction tree corresponding to the null set of variables, with $\mathcal{F}_\emptyset = \{\emptyset, \Omega\}$, and connecting it to all the individual junction trees resulting from each disconnected component. After propagation, $\phi_\emptyset^m(\Omega) = \underline{P}(E)$ and $\phi_\emptyset^Q(\emptyset) = \overline{P}(E)$. Unlike the case with pure probability, evidence $E$ in one component does, in general, influence the bounds of $\underline{P}(A|E)$ even when $A$ belongs to a disconnected component of the graph. Therefore, connecting the junction trees in the manner before propagation is mandatory.

**Theorem 4** *After a full propagation, $\mathbf{M}^{-1}[\phi_{\mathbf{C}}^m](A) = \underline{P}(A^\uparrow)$ for any $\mathbf{C} \in \mathcal{C}$ and $A \in \mathcal{F}_{\mathbf{C}}$, and $\mathbb{Q}^{-1}[\phi_{\mathbf{C}}^Q] = \overline{P}(A^\uparrow)$ for any $\mathbf{C} \in \mathcal{C}$ and $A \in \mathcal{F}_{\mathbf{C}}$.*

In other words, after a full propagation, the local potentials correctly encode the marginal lower probabilities.

### 4.2 Incorporation of Evidence

When it is known that the true situation is contained within a set $E \in \mathcal{F}$, we are interested in computing $\underline{P}(A^\uparrow|E)$ from the initially decomposed prior. The information, $E$, must therefore be incorporated into the local potentials. It is necessary to restrict $E$ to be a rectangular set on both $\mathcal{G}^m$ and $\mathcal{G}^Q$. We therefore assume that evidence is obtained incrementally, $E = E_1^\uparrow \cap E_2^\uparrow \cap ... \cap E_n^\uparrow$, such that each $E_i \in \mathcal{F}_{\mathbf{A}_i}$, $\mathbf{A}_i \subset \mathbf{C}_i^m \in \mathcal{C}^m$ and $\mathbf{A}_i \subset \mathbf{C}_i^Q \in \mathcal{C}^Q$. Each $E_i$ can therefore be successively incorporated, in any order, to condition on the total evidence $E$.

Proposition 1 demonstrates how evidence is incorporated into the potentials. It does not refer to local potentials, but it does naturally extend to local potentials as one might expect.

**Theorem 5** *Let $E \in \mathcal{F}_{\mathbf{C}}$. Suppose for $A \in \mathcal{F}$, $m(A) = \phi_{\mathbf{C}}(A_{\mathbf{C}}) \cdot f(A)$, where $\phi_{\mathbf{C}} : \mathcal{F}_{\mathbf{C}} \longrightarrow \Re$ and $f : \mathcal{F} \longrightarrow \Re$ are arbitrary. Suppose also that for all $A \in \mathcal{F}$,*

$$m'(A) = \begin{cases} m(A) & \text{if } A \subset E^\uparrow \\ 0 & \text{otherwise} \end{cases}$$

*Then $m'(A) = \phi'_{\mathbf{C}}(A_{\mathbf{C}}) \cdot f(A)$, where*

$$\phi'_{\mathbf{C}}(C) = \begin{cases} \phi_{\mathbf{C}}(C) & \text{if } C \subset E \\ 0 & \text{otherwise} \end{cases} \quad (9)$$

*Similarly, if $Q(A) = \phi_{\mathbf{C}}(A_{\mathbf{C}}) \cdot f(A)$, and*

$$Q'(A) = \begin{cases} Q(A) & \text{if } A \subset E^\uparrow \\ 0 & \text{otherwise} \end{cases}$$

*then $Q'(A) = \phi'_{\mathbf{C}}(A_{\mathbf{C}}) \cdot f(A)$ where $\phi'$ is given by (9).*

Theorem 5 says that to incorporate evidence $E \in \mathcal{F}_{\mathbf{A}}$, it is only necessary to find one node $\mathbf{C} \supset \mathbf{A}$ in each junction tree and adjust the local potential for $\mathbf{C}$. This is done by zeroing out all local potential assignments for sets that are not subsets of $E$. Evidence can thus be locally incorporated. Note that it is not necessary for the local potentials to be pairwise consistent — i.e., evidence can be incorporated at any time, before or after propagation.

There are two possible singularities that can arise when evidence is incorporated. If $\overline{P}(E) = \underline{P}(E) = 0$, then the event $E$ is impossible and entirely in conflict with the prior lower probability assignment. This case is quickly recognized during the propagation of $\phi^Q$ when $x/0, x \neq 0$ is encountered, or when some local $\phi^Q(\emptyset)$ potential becomes zero. In this case it should be reported that a logical contradiction has been encountered. The same singularity can, of course, occur with pure probability. A second singularity is more subtle and less fatal, and occurs when $\overline{P}(E) > \underline{P}(E) = 0$. In this case, $E$ being impossible is consistent with the prior, but not necessarily so, so there is no logical contradiction. It is entirely legitimate for this situation to occur. It can be detected during the propagation of $\phi^m$ when $x/0, x \neq 0$ is encountered or when a local $m$ potential, $\phi^m$, becomes zero everywhere during propagation, but when the same singularity does not occur in $\phi^Q$. When this happens, $\underline{P}(A|E) = 1$ when $A \subset E$ and zero otherwise ([9]). Whether $A^\uparrow \in E$ can be readily obtained from the propagated $\phi_{\mathbf{A}}^Q$ from local information, so the propagation of $\phi^Q$ should be completed. It is an inherent disadvantage of lower probability, and not of its graphical representation, that all grades of uncertainty are lost whenever a plausibly impossible conditioning event is encountered.

## 5 Conclusion

The full propagation algorithm for 2-monotone lower probabilities can be summarized as follows:

1. Obtain undirected dependency graphs for $\mathbf{M}[\underline{P}]$ and $\mathbb{Q}[\overline{P}]$. These may (optionally) be different graphs.

2. Triangulate the graphs.

3. Extract junction trees for each graph.

4. Initialize junction tree potentials to encode the prior lower probabilities according to (8). One junction tree encodes $\mathbf{M}[\underline{P}]$, the other $\mathbb{Q}[\overline{P}]$.

5. For each local piece of evidence, find one node in each junction tree that discerns that evidence. Update these local potentials according to (9).

6. Propagate the potentials.

7. To obtain $\underline{P}(A^\uparrow|E)$ or $\overline{P}(A^\uparrow|E)$ for some local $A \in \mathcal{F}_{\mathbf{A}}$, read off $\underline{P}(A^\uparrow \cap E)$ directly from a local node in the first junction tree, and $\overline{P}(A^\uparrow \cap E)$ directly from a local node in the second junction tree. Use (2) to obtain $\underline{P}(A^\uparrow|E)$ or $\overline{P}(A^\uparrow|E)$.

Although $\underline{P}(A|E)$ appears resistant to exact decomposition and propagation algorithms, it can be propagated by breaking it into two components, $\underline{P}(A \cap E)$ and $\overline{P}(A \cap E)$, each of which is amenable to propagation. This central observation is from [9].

Because both $\mathbf{M}[\underline{P}]$ and $\mathbf{Q}[\overline{P}]$ must be decomposable, it is clear that the structural requirements for decomposability of lower probabilities are rather strict. This is a serious limitation. On the other hand, it is not entirely surprising that decomposability requirements associated with both the lower bounds as well as with the upper bounds might exist. The fact that they can be decomposed in different ways may help to ease this extra structural requirement somewhat.

It is important to develop frameworks in which decomposable lower probabilities can be naturally expressed and constructed from smaller bits and pieces. The double Markov requirement makes this endeavor challenging but even more important. It may also be of interest to develop approximation methods for loosening bounds in order to achieve the double Markov condition without adding an excessive number of extra edges.

Conditional independence properties associated with decompositions of pure probability have been heavily studied [17, 30, 36]. However, one should take caution in making similar interpretations within a lower probability framework ([13]), the same intuitions do not always transfer. It can be shown that there are severe limitations in the ability of a lower probability representation to express epistemological independence, the idea that knowledge about one event should not influence the bounds for an independent event ([10]). This has significant ramifications on the interpretation of lower probabilities. The study of lower probability interpretation is very important, particularly in the context of the decomposable graphical framework where little previous attention has been focused.

The ability to decompose and propagate lower and upper probabilities offers significant potential for expanding their many uses to larger applications. Further improvements in the tractability are severely needed and provide many important areas for future research. A straightforward non-parametric potential representation requires arrays of size exponential in the number of joint assignments to the variables in a clique, since each subset is assigned a potential value. Without further developments, this limits the propagation algorithm to graphs with very small cliques. The study of conjugate parametric representations for lower probabilities has been almost entirely overlooked, but could be very valuable towards these ends. Even more promising is the study of sparse representations for Möbius and commonality assignments in the context of propagation, i.e., where most Möbius assignments are zero ([9]). Such an approach might allow very large clique sizes provided that the potentials themselves are very sparse.

## Acknowledgements

I thank Fabio Cozman for comments on a draft of this paper. The author was supported by NASA-JPL Grant NGT-51039 and by NASA grant NAGW-1175. The views contained within are those of the author and do not represent the official policies, either expressed or implied, of the U.S. government.